\documentclass[runningheads,orivec]{llncs}
\usepackage[T1]{fontenc}
\usepackage{graphicx}
\usepackage{cite}
\usepackage{xspace}
\usepackage{url}
\usepackage[hidelinks]{hyperref}
\usepackage{booktabs}
\usepackage{array}
\usepackage{amsmath}
\usepackage{amssymb}
\usepackage{adjustbox}
\usepackage{microtype}  % For some older microtype package versions this triggers a \showhyphens warning that you cannot do anything about (internal package bug)

\AtBeginDocument{%
\setlength{\abovedisplayskip}{6pt plus 2pt minus 2pt}%
\setlength{\belowdisplayskip}{6pt plus 2pt minus 2pt}%
\setlength{\abovedisplayshortskip}{4pt plus 2pt minus 2pt}%
\setlength{\belowdisplayshortskip}{4pt plus 2pt minus 2pt}%
}

\renewcommand{\figureautorefname}{Fig.\@\!}
\newcommand{\seclabel}[1]{\label{sec:#1}}
\newcommand{\figlabel}[1]{\label{fig:#1}}
\newcommand{\tablabel}[1]{\label{tab:#1}}

\newcommand{\secref}[1]{\autoref{sec:#1}\xspace}
\newcommand{\figref}[1]{\autoref{fig:#1}\xspace}
\newcommand{\tabref}[1]{\autoref{tab:#1}\xspace}

\newcommand{\figrefrange}[2]{\figureautorefname~\ref{fig:#1}--\ref{fig:#2}}

\newcommand{\ie}{i.e.\@\xspace}
\newcommand{\eg}{e.g.\@\xspace}
\usepackage{color}

\begin{document}
\title{Identifying the Unknown: Prompt-Free Open Vocabulary Anomaly Recognition for Robot-Object Interaction}
\titlerunning{Identifying the Unknown}
\author{Philipp Allgeuer \and Jan-Gerrit Habekost \and Stefan Wermter\thanks{Work gratefully funded by German Research Foundation (DFG) project 551629603 (LUMO) and Horizon Europe MSCA grant agreement 101072488 (TRAIL).\\
The authors have no competing interests to declare for this work.}}
\authorrunning{P. Allgeuer et al.}
\institute{Knowledge Technology, Dept. of Informatics, University of Hamburg, Germany\\
\email{\{philipp.allgeuer,jan-gerrit.habekost,stefan.wermter\}@uni-hamburg.de}}
\maketitle
\vspace{-3.5ex}
\begin{abstract}
Robots operating in real-world environments must in general be able to recognize previously unseen objects. As robotic systems move toward open-world autonomy, there is a growing, yet largely unmet, need for open vocabulary object detectors that are prompt-free and efficient enough for continuous deployment. We present AnomNOVIC, a two-stage known-workspace framework that combines a masked autoencoder (MAE) trained for anomaly detection, with NOVIC, a powerful real-time prompt-free open vocabulary image classifier. The MAE produces generic object-agnostic bounding boxes, allowing NOVIC to classify salient image regions without requiring a predefined candidate class list. We evaluate AnomNOVIC against strong open vocabulary baselines in a tabletop robot-object environment featuring the NICOL humanoid robot, reaching 47.1\%\! AP / 57.5\%\! AP\textsubscript{50} for prompt-free recognition, and 59.0\%\! AP / 72.5\%\! AP\textsubscript{50} if class candidates are provided. Across additional datasets, including an in-the-wild test set with 48 unique objects, AnomNOVIC reaches up to 82.6\% prompt-free detection and classification accuracy. These results significantly surpass all tested open vocabulary baselines, including YOLO-World-v2, OWLv2, and YOLOE.
\vspace*{-2ex}
\end{abstract}

%%%%%%%%%%%%%%%%%%%%%%%%%%%%%%%%%%%%%%%%%%%%%%%%%%%%%%%%%%%%%%%%%%%%%%%%%%%%%%%%
\vspace{-6ex}
\section{Introduction}
\seclabel{introduction}
\vspace{-2ex}

Robotic agents driven by Large Language Models (LLMs) are transforming human-robot interaction, enabling natural dialogue, reasoning, and task planning. These high-level capabilities, however, often remain bottlenecked by a robot's real-time visual perception. Open vocabulary object detectors are a promising approach to this challenge, as they enable zero-shot discovery and identification of previously unseen objects encountered at runtime, especially in unstructured or dynamic environments. Existing open vocabulary object detectors, however, come at varying levels of true `openness'. Many state-of-the-art detectors, including ViLD \cite{gu2022_vild}, YOLO-World \cite{Cheng2024YOLOWorld}, and OWLv2 \cite{minderer2023scaling}, require a list of candidate object classes at inference time (`prompted' models), which is a strong constraint that severely limits their practical utility. If a robot can only detect objects it has explicitly been instructed to look for, then it cannot actually recognize any novel objects in practice, undermining the core supposed advantage of open vocabulary detection. Prompted models also tend to fail ungracefully in the presence of novel objects, frequently producing false positives. Open vocabulary models that do not require any list of candidate classes are referred to as \emph{prompt-free} \cite{wang2025yoloerealtimeseeing}.

A second common limitation of state-of-the-art open vocabulary detectors is that many are trained or fine-tuned on datasets of limited concept coverage. Even if a model is technically `open vocabulary', it will still fail to detect concepts that are insufficiently represented in its training data. Thus, if a training dataset contains only a few thousand unique object concepts, any resulting model will in practice have a rather limited active vocabulary at inference time. This issue is exacerbated by the need for substantial examples per concept to facilitate effective learning and generalization. As a reference, the RAM model \cite{Zhang_2024_CVPR} constructed its vocabulary by parsing and expanding upon 14M training captions, ultimately arriving at only 4585 tags---including many non-object terms, and only after manually adding many missing terms ($\approx$\,1K) from known classification datasets. If we compare this to the open vocabulary prompted YOLO-World \cite{Cheng2024YOLOWorld} and prompt-free YOLOE \cite{wang2025yoloerealtimeseeing} models, both trained on under 1.6M images, we can see that the `openness' they can achieve is strongly limited.

Among current open vocabulary models, our previous work NOVIC \cite{allgeuer_novic_2025} is rather unique in being capable of real-time prompt-free image classification over a near-unconstrained taxonomy of object nouns ($\approx$\,43K). However, while NOVIC excels at assigning fine-grained free-form classifications to images as a whole, it can neither classify nor localize multiple individual objects within an image, which is crucial for real applications. In this work we present AnomNOVIC, an integration of NOVIC's prompt-free classification capabilities into an open vocabulary object detection pipeline, and apply it to the tabletop scenario of the NICOL robot (see \figref{pipeline_overview}). A key challenge is how to identify suitable regions of interest (RoIs) without introducing class or vocabulary biases. AnomNOVIC addresses this by training a transformer-based anomaly detection masked autoencoder (MAE) \cite{MaskedAutoencoders2021} to model the robot's \emph{object-free} operational workspace as a normative baseline, and flagging reconstruction anomalies at inference time as object region proposals. Additional anomaly mask and table mask output channels further enhance the quality of these proposals. The final obtained RoIs are cropped from the input images and passed to NOVIC, which assigns accurate fine-grained free-form labels. By decoupling object localization from classification, AnomNOVIC leverages the axiomatic notion of ``anything that does not belong'' to discover objects of any shape, texture, or novelty in an unbiased manner.

%%%%%%%%%%%%%%%%%%%%%%%%%%%%%%%%%%%%%%%%%%%%%%%%%%%%%%%%%%%%%%%%%%%%%%%%%%%%%%%%
\vspace{-2.5ex}
\section{Related Work}
\seclabel{related_work}
\vspace{-2ex}

\begin{figure}[!t]
\parbox{\linewidth}{%
\centering%
\adjustbox{valign=c}{\includegraphics[width=0.46\linewidth, trim=0px 15px 0px 0px, clip]{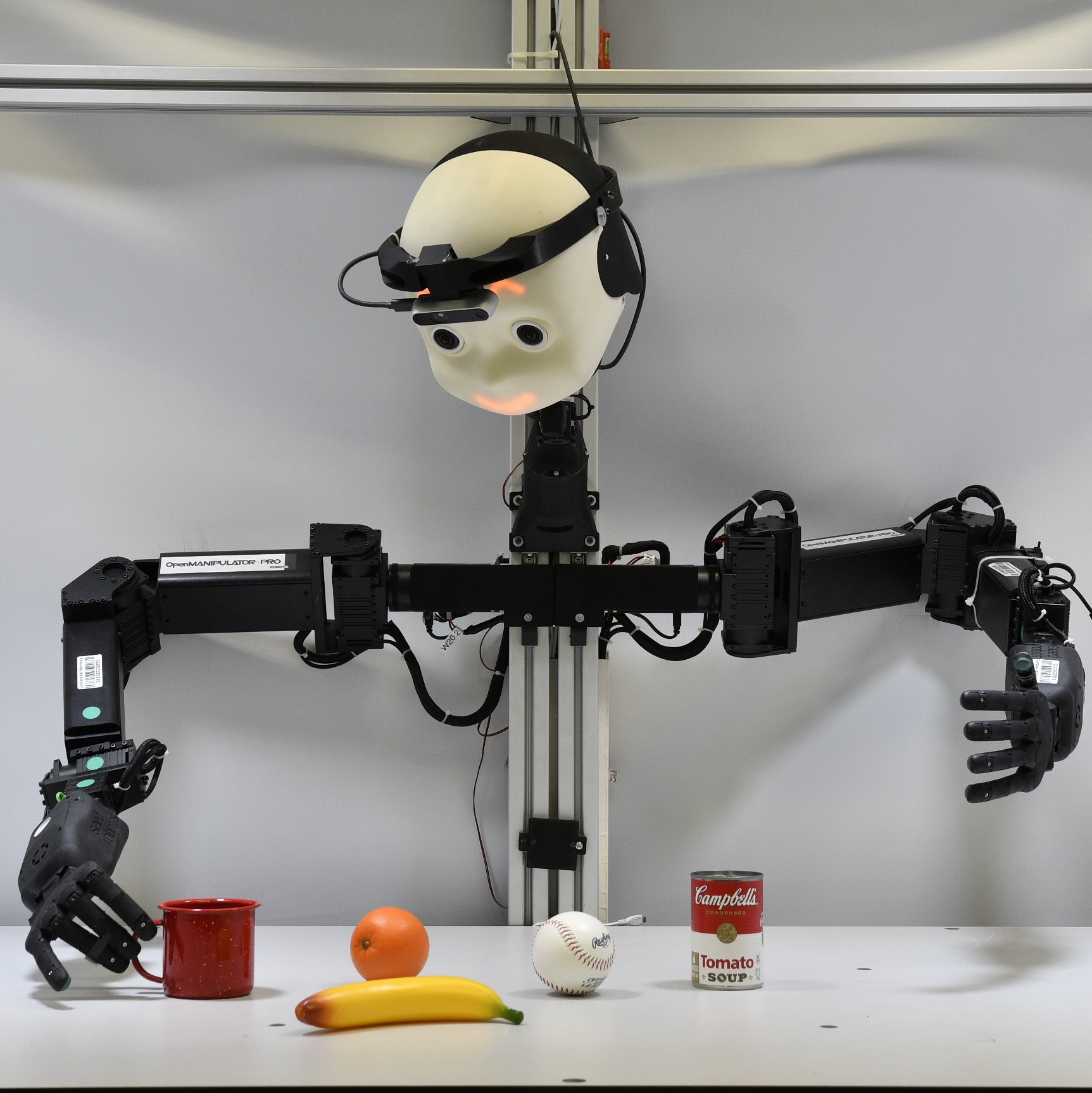}}\hfill%
\adjustbox{valign=c}{\includegraphics[width=0.5\linewidth]{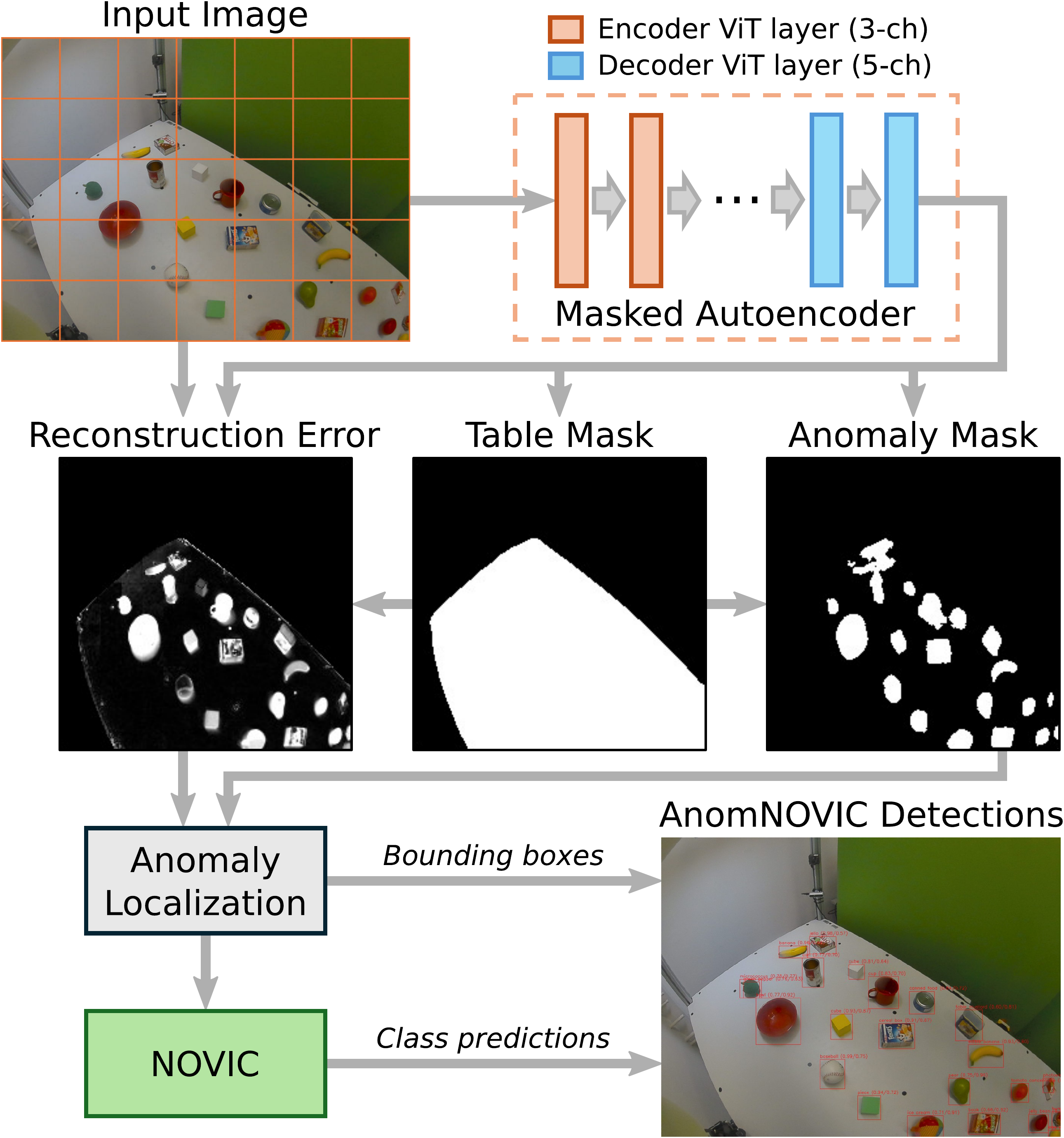}}%
}%
\vspace*{-2ex}
\caption{\textbf{Left:} The LLM-powered NICOL is used in a human-robot-object interaction scenario requiring open vocabulary object detection. \textbf{Right:} A two-stage anomaly recognition model, AnomNOVIC, unites an anomaly detection masked autoencoder (MAE) with a strong prompt-free open vocabulary image classification model (NOVIC).}
\figlabel{pipeline_overview}
\vspace*{-4ex}
\end{figure}

%===============================================================================
\subsection{Open Vocabulary Object Detection}
\seclabel{related_ovod}
\vspace{-1.5ex}

Open vocabulary object detectors can be broadly divided into prompted and prompt-free approaches. ViLD \cite{gu2022_vild} was among the first to use pretrained vision-language models for open detection, passing prompt-templated class names through a CLIP text encoder \cite{pmlr-v139-radford21a} to form classification weights for region proposals from a distilled class-agnostic backbone. Other prompted detectors such as RegionCLIP \cite{zhong2022regionclip}, DetCLIPv3 \cite{Yao_2024_CVPR}, and Grounding DINO \cite{Liu_grounding_dino_2025} extend this idea by aligning regional features with text embeddings and/or unifying detection and phrase grounding via multimodal pretraining. For fast inference, YOLO-World \cite{Cheng2024YOLOWorld} adapts the efficient single-stage anchor-free YOLOv8 backbone and adds a reparameterizable vision-language path aggregation network. Models like OVR-CNN \cite{Zareian_2021_CVPR} and OWL-ViT \cite{minderer2022simple} fine-tune detection heads on datasets of restricted diversity and rely on contrastive image-text pretraining for implicit generalization. OWLv2 \cite{minderer2023scaling} scales this up by self-training on web-scale image-text data pseudo-labeled by OWL-ViT, but remains non-real-time and still requires explicit class prompts at inference. The self-training mainly consolidates concepts OWL-ViT can already recognize, rather than effectively learning genuinely novel categories. While these methods can to some degree recognize arbitrary classes specified at inference time, they still require a user-supplied class vocabulary, and their performance depends heavily on the scope of the underlying training data.

YOLOE \cite{wang2025yoloerealtimeseeing} seeks to alleviate the first of these constraints by supporting a ``prompt-free'' mode. This prompt-free mode frames open vocabulary classification as a retrieval task based on a fixed mid-sized reference vocabulary. A special prompt embedding selects generic class-agnostic object proposals, and the corresponding object embeddings are matched against the complete built-in vocabulary for category retrieval. This fixed vocabulary is exactly the limited 4585 RAM tags \cite{Zhang_2024_CVPR} discussed in \secref{introduction}. The use of this limited fixed vocabulary, combined with training on datasets of limited concept coverage, imposes an upper bound on YOLOE's true `openness'.
NOVIC \cite{allgeuer_novic_2025} is a recent model that achieves prompt-free open vocabulary image classification by leveraging the knowledge of a pretrained CLIP model. It has dictionary-level coverage of the English language, with its classification decoder trained exclusively on content-controllable synthetic text datasets. The model, however, cannot localize objects or classify multiple of them per image. NOVIC serves as the foundation for the AnomNOVIC model introduced in this work, which addresses these limitations.

%===============================================================================
\vspace{-3ex}
\subsection{Anomaly Detection}
\seclabel{related_anomaly_detection}
\vspace{-1.5ex}

Unsupervised visual anomaly detection is traditionally framed as a reconstruction task, where a model trained solely on normal samples attempts to reconstruct the input, and flags high-error regions. Vision Transformers (ViTs) have recently been used to address locality limitations of CNNs in anomaly detection. AnoViT \cite{anovit} combines a ViT encoder with a CNN decoder to capture global patch relations and improve localization on MVTec AD \cite{Bergmann2021}. ViTALnet \cite{vitalnet} extends this with a pyramidal architecture and integrated global attention, while UTRAD \cite{CHEN2022} uses a U-Transformer with skip connections to handle structural and texture anomalies across scales. MAEDAY \cite{SCHWARTZ2024} applies an ImageNet-pretrained masked autoencoder directly at test time, where failed reconstruction reveals unseen objects without domain-specific training. Some methods amplify reconstruction loss by synthetically injecting defects during training, such as DRAEM \cite{Zavrtanik_2021_ICCV}, which pastes foreign textures onto normal images and jointly learns inpainting and segmentation. In our work, we train a masked autoencoder from scratch on images of the robot's empty workspace, with additional augmentation via object pasting. Two supplementary output channels explicitly predict anomaly and table masks alongside the reconstruction to assist anomaly localization.

%%%%%%%%%%%%%%%%%%%%%%%%%%%%%%%%%%%%%%%%%%%%%%%%%%%%%%%%%%%%%%%%%%%%%%%%%%%%%%%%
\vspace{-2.5ex}
\section{Approach}
\seclabel{approach}
\vspace{-2ex}

The proposed two-stage AnomNOVIC pipeline, as applied to the NICOL scenario, is shown in \figref{pipeline_overview}. In the first stage, a ViT masked autoencoder (MAE) is trained solely on object-free snapshots of the robot's workspace, allowing it to learn to reconstruct the scene. This includes explicit estimation of the table area in the form of a binary mask. An object pasting augmentation strategy, \ie the pasting of numerous synthetic anomalies into each input image, is used to train the MAE to produce a further mask output that explicitly highlights anomalous regions. The target outputs of the autoencoder do not include the pasted anomalies, so at inference time, when objects are on the table, the generated reconstructions omit them, causing significant reconstruction errors in those regions. Together with the anomaly and table masks, this is used to resolve bounding boxes for all anomalies in the image, effectively combining the strengths of the reconstruction and anomaly masks for a more reliable output. As \emph{any} object not present in the clean workspace training images is uniformly treated as an anomaly, the resulting detections are highly class-agnostic and unbiased.
Given these anomaly bounding boxes from the first stage, corresponding but slightly enlarged anomaly-centered crops are taken from the original input image and used for batch inference with NOVIC---a real-time open vocabulary image classifier capable of generating fine-grained free-form object noun labels without requiring any class candidates \cite{allgeuer_novic_2025}. The final confidence score for each detection combines the NOVIC prediction probability with a measure of the average confidence of the corresponding anomaly mask or reconstruction error, further adjusted by additional factors as detailed in \secref{anomaly_localization}. To promote reproducibility and comparability, we use a pretrained NOVIC checkpoint for all main results. However, we also train our own NOVIC variants from scratch on a tuned version of its training dataset (see \secref{novic}).

%===============================================================================
\vspace{-2.5ex}
\subsection{Scenario and Dataset Collection}
\seclabel{scenario_dataset_collection}
\vspace{-1ex}

NICOL (see \figref{pipeline_overview}) is an LLM-powered semi-humanoid research platform that serves as a testbed for neural approaches in manipulation and human-robot interaction. It is a tabletop robot comprising only an upper torso, and includes a 2-DoF pan-tilt head with a camera in each eye, enabling dynamic observation of the user and workspace, and perception of any objects on the table. Given object bounding box detections paired with open vocabulary textual class labels, NICOL can use its underlying LLM agent to discuss, show, move, and manipulate the objects on the table at the user's verbal request. This is a crucial capability for many robot experiments and tasks, directly taking advantage of the prompt-free nature of AnomNOVIC to allow seamless interaction with totally novel objects at runtime. Two train/test dataset pairs were collected for the NICOL workspace (see \figref{train_test_data}), one with and one without known workspace distractor objects.

%-------------------------------------------------------------------------------
\vspace{-3ex}
\subsubsection{Training Dataset}
\seclabel{training_dataset}

\begin{figure}[!t]
\linethickness{0.8pt}%
\centering%
\frame{\parbox{\linewidth}{\centering%
\includegraphics[width=0.3333\linewidth]{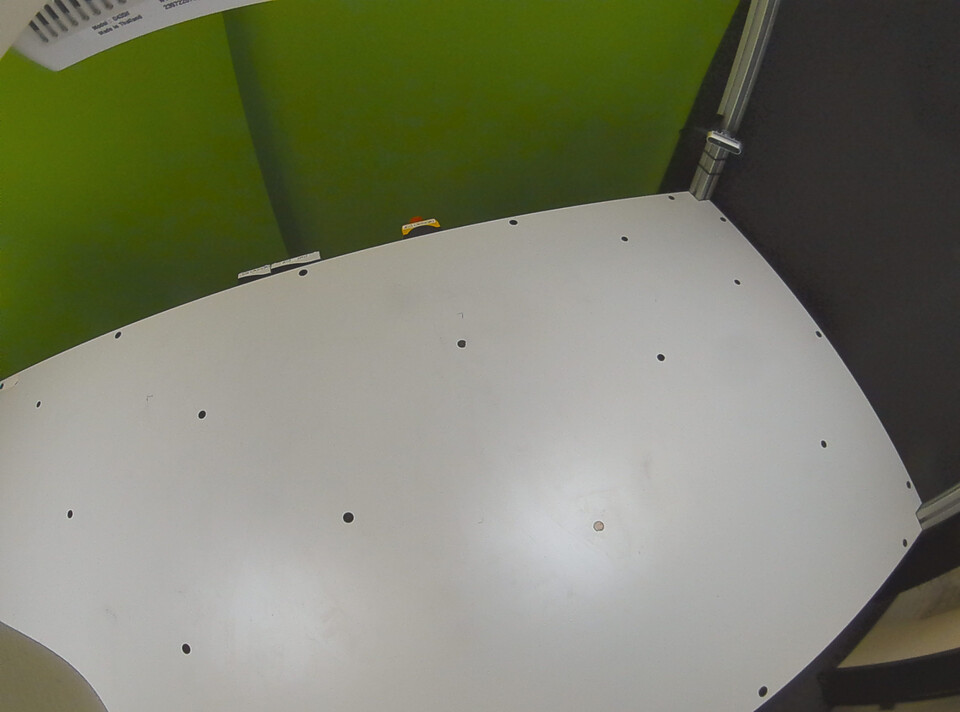}%
\includegraphics[width=0.3333\linewidth]{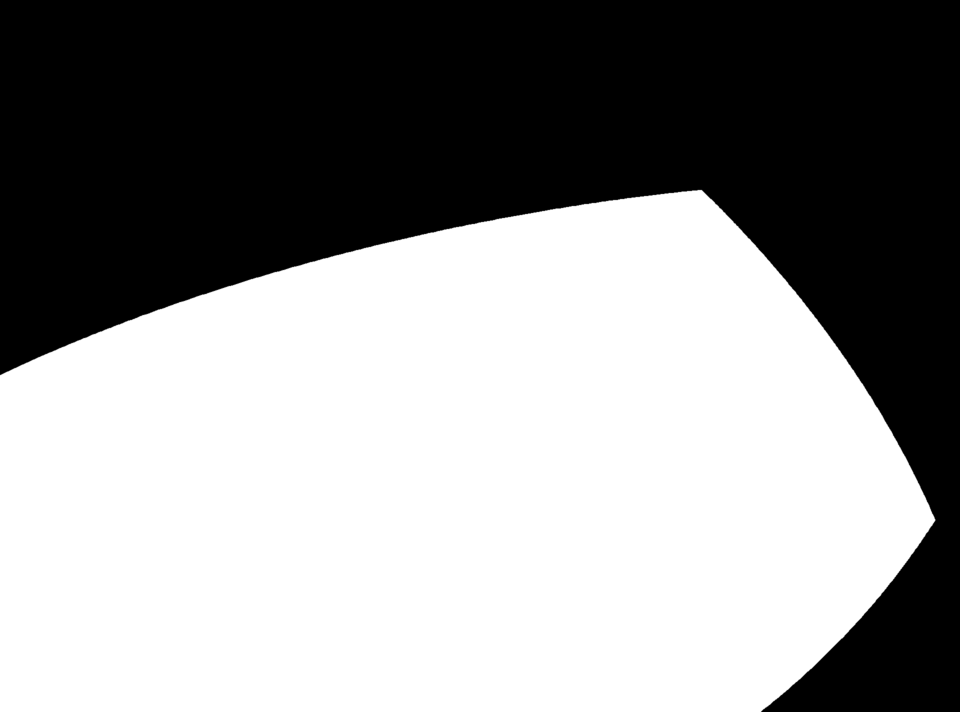}%
\includegraphics[width=0.3333\linewidth]{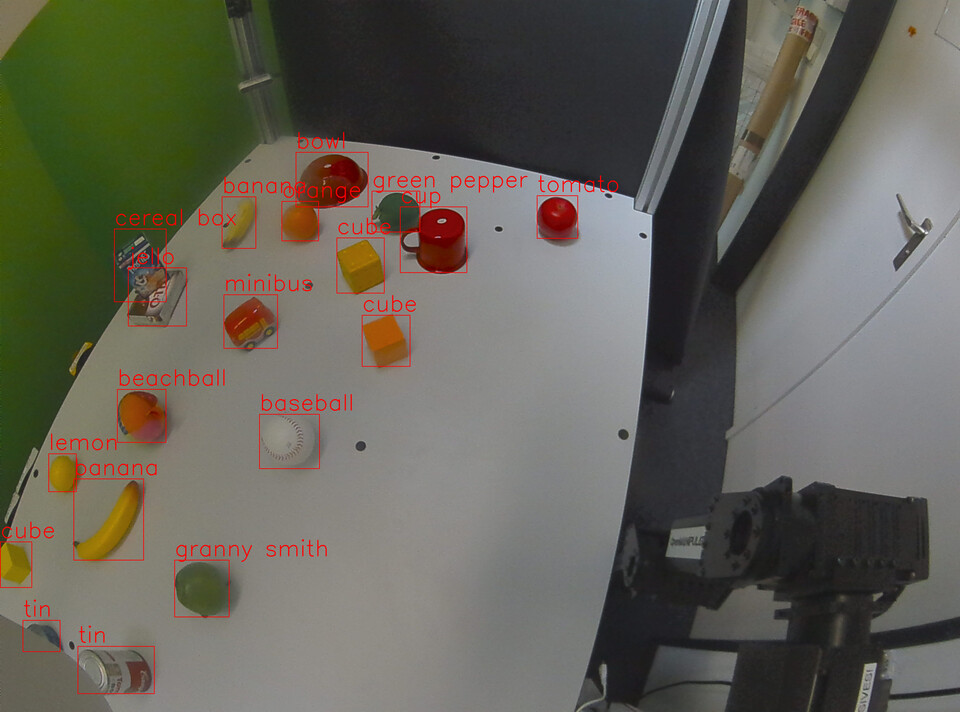}\\\nointerlineskip
\includegraphics[width=0.3333\linewidth]{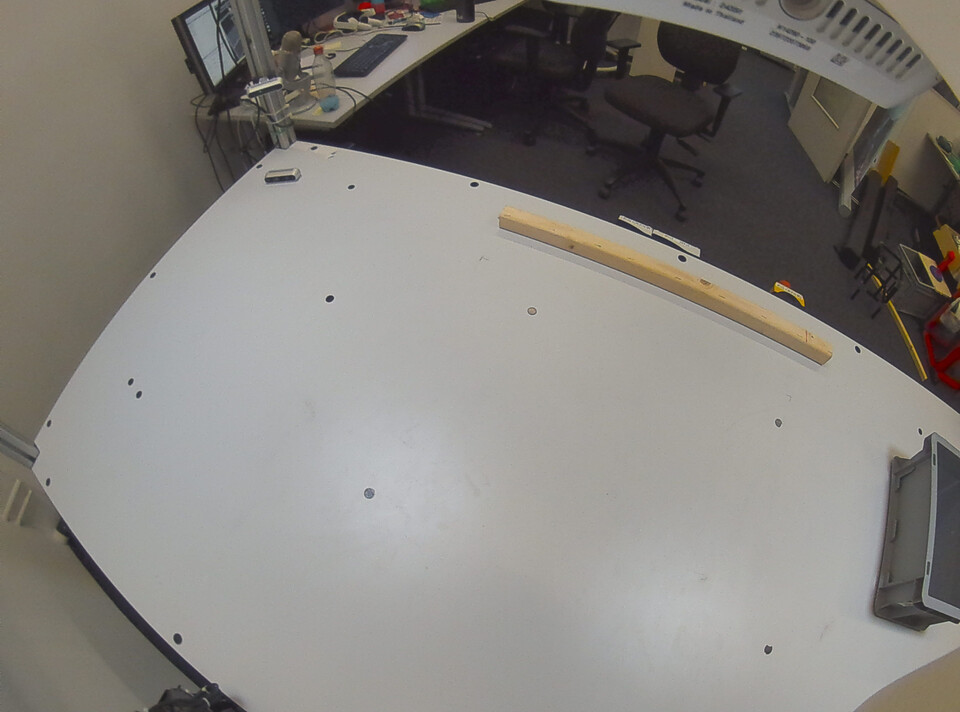}%
\includegraphics[width=0.3333\linewidth]{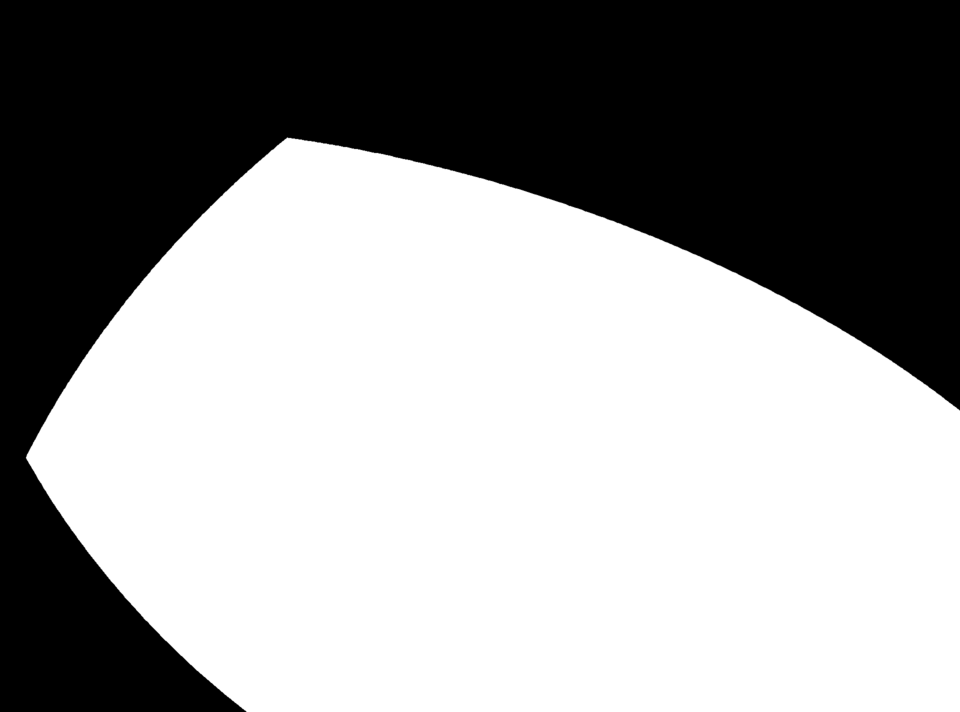}%
\includegraphics[width=0.3333\linewidth]{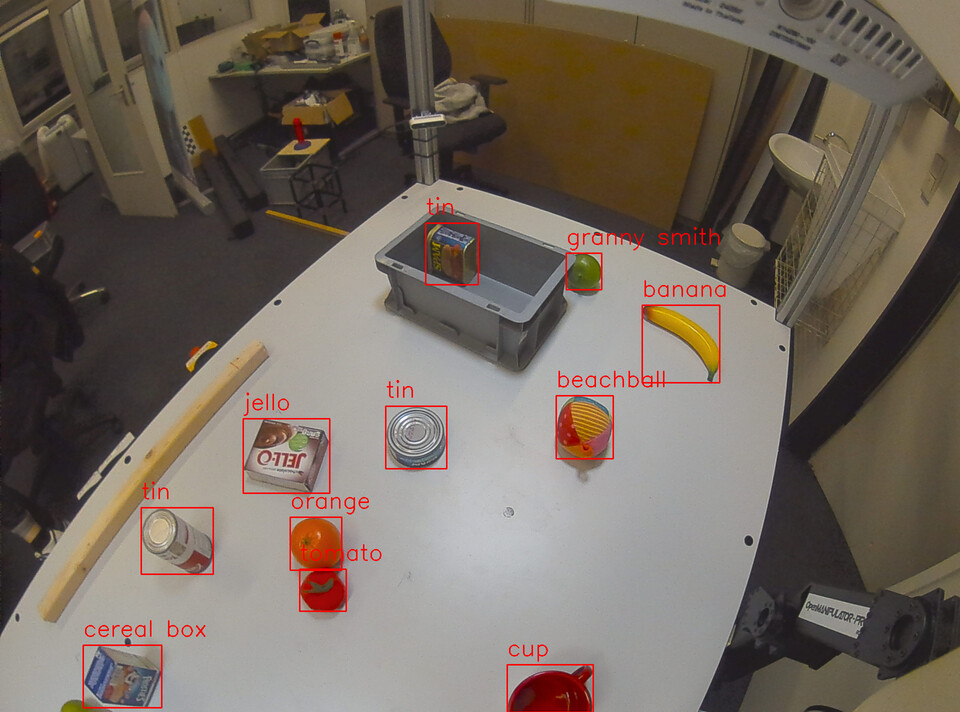}%
}}%
\vspace*{-2ex}
\caption{Examples from our training (left, center) and test (right) datasets, with (bottom) and without (top) distractor objects. The training dataset only contains clean images of the workspace and corresponding table masks (making it quick and easy to collect), while the test dataset contains anomalous images with many (often overlapping) objects.}
\figlabel{train_test_data}
\vspace*{-4ex}
\end{figure}

To collect training data, NICOL's head was simply rotated through a grid of head poses while capturing snapshots of its empty tabletop workspace (only possibly including distractor objects) under varying lighting, background conditions, and arm positions. This fast, simple, and annotation-free procedure yielded 7500 clean normal samples per dataset. Using measured joint states together with a calibrated camera model \cite{gajdosech2025}, binary table masks were automatically generated for each sample without any manual annotation. For this, the tabletop boundary was sampled at regular intervals, projected into the image frame, and converted into a near pixel-perfect mask via Delaunay triangulation.

%-------------------------------------------------------------------------------
\vspace{-3ex}
\subsubsection{Test Datasets}
\seclabel{test_datasets}

Test data was collected by repeatedly arranging random subsets of a pool of 28 physical objects (corresponding to 17 unique classes) on the workspace in front of the NICOL and recording each scene from 5 unique head poses. For the plain (\ie no distractors) dataset, this resulted in 117 scenes and 585 RGB test images. Subsequent manual annotation yielded 13578 tight shadow-free ground truth bounding boxes. The distractors dataset was collected analogously and has similar statistics. A supplementary \emph{Wild} test dataset was also similarly recorded (without distractors), and comprises 38 images featuring 48 unique and rarer in-the-wild objects (\eg \emph{karabiner}, \emph{audio amplifier}), with 5--11 objects per image and 310 object instances in total. This test dataset serves to evaluate and demonstrate the true openness of the AnomNOVIC pipeline.

The test sets are quite challenging for a number of reasons, including due to the presence of specular reflections, shadows, partial occlusions, sensor noise, imperfect image sharpness, uncommon object placements (\eg upside-down), and fisheye lens distortion, which frequently causes unusual object perspectives. The inclusion of less typical object variants, \eg empty lidless tins, plastic/felt produce, and similarly unconventional items, further complicates open-set recognition. Image crops illustrating these compounded challenges are shown in \figref{test_object_crops}. The test sets are also challenging due to their generally small object sizes, with a median size (geometric mean of width and height) of 99 pixels and a minimum size of only 30 pixels at the native 1920$\times$1440 resolution. When preprocessed to the nominal 224px working resolution of the MAE, the object sizes drop by a further factor of $\approx$\,7. For a closed-set detector, this may not pose a major issue, as simple color or shape cues can often easily disambiguate a fixed set of known classes, but for open-set detectors this is a significant challenge. A truly open vocabulary, \ie prompt-free, model must consider virtually the entire English vocabulary as classification options, and therefore indispensably needs enough visual detail to make a confident prediction. For example, to correctly classify a cardboard box as \emph{jello}, the model must either uniquely recognize the box design among all possible existing products, or at minimum be able to pseudo-read the stylized `Jell-O' text from varying perspectives, orientations, and scales.

%===============================================================================
\vspace{-2.5ex}
\subsection{Masked Autoencoder for Anomaly Detection}
\seclabel{masked_autoencoder}
\vspace{-1.5ex}

We build on the masked autoencoder (MAE) \cite{MaskedAutoencoders2021}, and adapt it for anomaly detection for the purpose of class-independent object detection. We select the ViT-B/16 architecture, operating at 224$\times$224 resolution, and shorten the encoder to 6 layers while keeping all 8 decoder layers. A smaller network is beneficial, as the NICOL workspace exhibits less appearance variation than large-scale datasets like ImageNet-1K, improving both training efficiency and inference speed ($\approx$\,9\,ms on an RTX A6000). Patch masking remains crucial in compelling the autoencoder to infer higher-level regional structure rather than just relying on local image statistics, but mask ratios sampled uniformly per batch from 0\% to 30\% were empirically determined to perform better than the original fixed 75\%. After 1600 of the 2000 training epochs, the mask ratio is also set to 0\%, effectively finetuning the model for inference with 0\% masking, and together with cosine learning rate decay helping the model converge on high quality reconstructions. Throughout training, we use several augmentations to improve generalization, including color jitter, random perspective distortions, rotations of up to $30^\circ$, and random resize-cropping down to 224px.

The original MAE decoder is extended to predict three aligned outputs instead of one---the reconstructed RGB image, a table mask, and an anomaly mask---resulting in a 5-channel output tensor. The table and anomaly mask channels use sigmoid activations to produce dense probability maps that highlight the table area (irrespective of occlusions) and any anomalous regions, respectively. The anomaly mask is supervised via an object pasting augmentation in which 2--25 virtual anomalies from a pool of 90 RGBA object images are pasted into the table area of each input image with random sizes, locations, rotations, and color jitter. These pasted objects are omitted from the target reconstruction image so that the MAE learns to identify and remove anomalies, and their pixels define the target anomaly mask. At inference, a denoising circular 9$\times$9 mean filter is applied to the predicted table mask prior to thresholding, and the resulting \emph{binary table mask} is used to mask the anomaly probability map, yielding (after thresholding) a \emph{binary anomaly mask} that highlights anomalous regions. Example predictions, targets, and the object pasting augmentation strategy are shown in \figrefrange{pipeline_overview}{test_object_crops}.

\begin{figure}[!t]
\linethickness{0.8pt}%
\centering%
\frame{\parbox{0.4334\linewidth}{\centering%
\includegraphics[width=0.3333\linewidth, height=0.3333\linewidth]{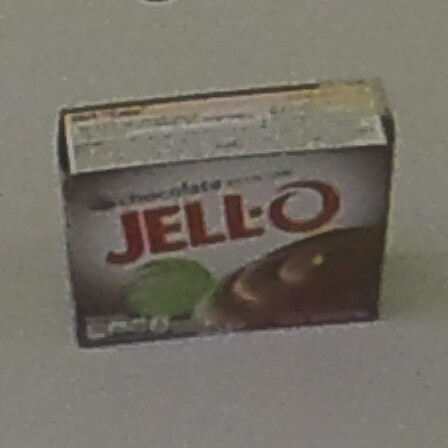}%
\includegraphics[width=0.3333\linewidth, height=0.3333\linewidth]{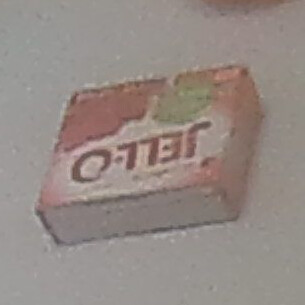}%
\includegraphics[width=0.3333\linewidth, height=0.3333\linewidth]{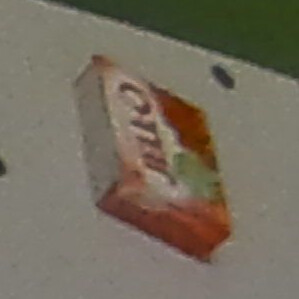}\\[-1pt]\nointerlineskip
\includegraphics[width=0.3333\linewidth, height=0.3333\linewidth]{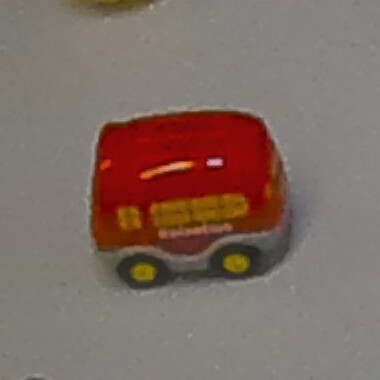}%
\includegraphics[width=0.3333\linewidth, height=0.3333\linewidth]{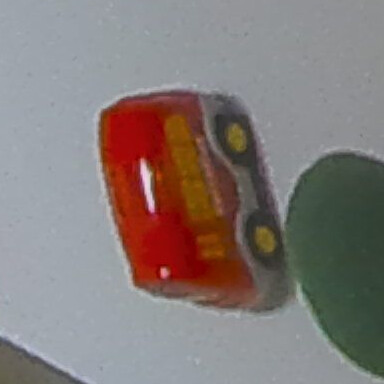}%
\includegraphics[width=0.3333\linewidth, height=0.3333\linewidth]{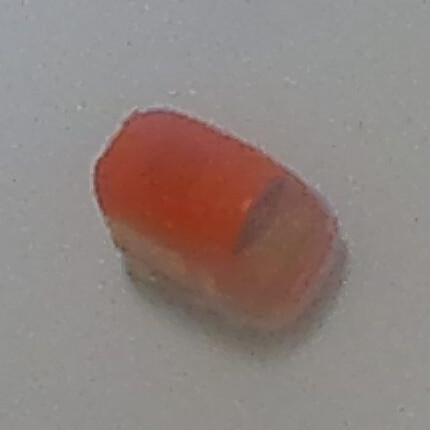}\\[-1pt]\nointerlineskip
\includegraphics[width=0.3333\linewidth, height=0.3333\linewidth]{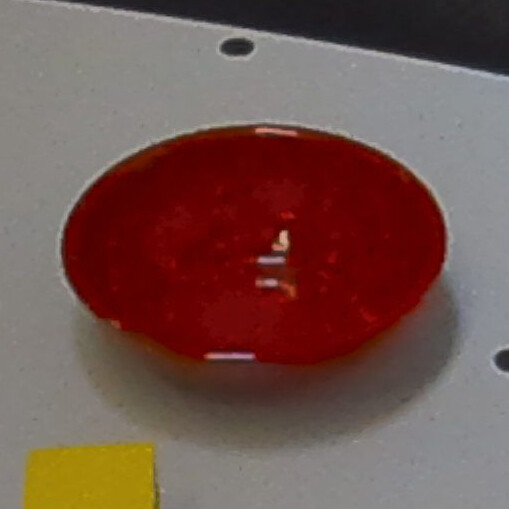}%
\includegraphics[width=0.3333\linewidth, height=0.3333\linewidth]{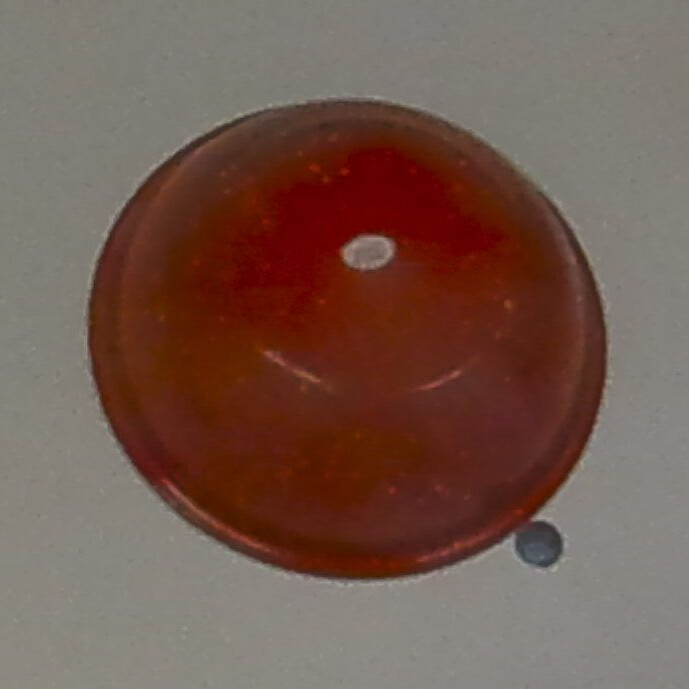}%
\includegraphics[width=0.3333\linewidth, height=0.3333\linewidth]{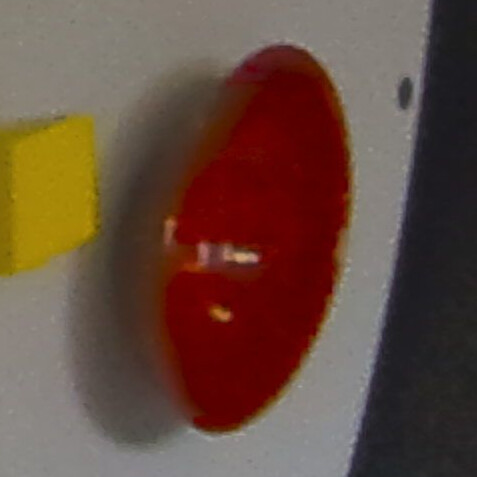}\\[-1pt]\nointerlineskip
\includegraphics[width=0.3333\linewidth, height=0.3333\linewidth]{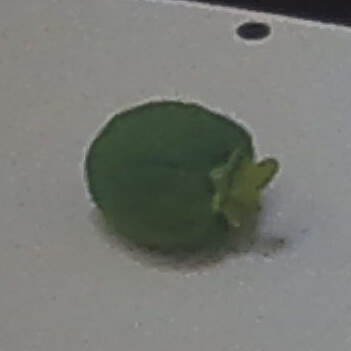}%
\includegraphics[width=0.3333\linewidth, height=0.3333\linewidth]{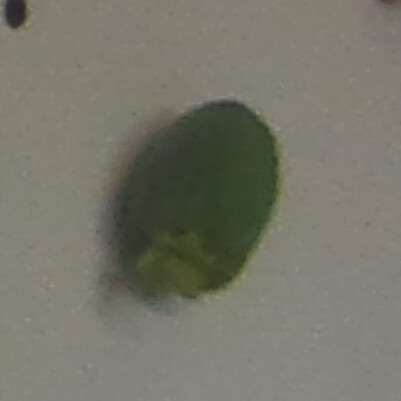}%
\includegraphics[width=0.3333\linewidth, height=0.3333\linewidth]{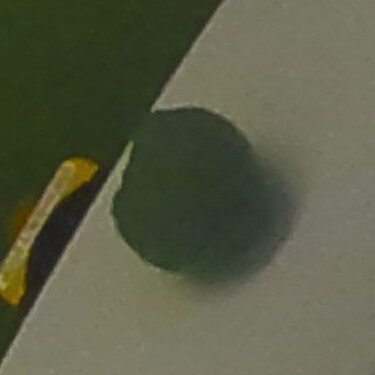}\vspace*{-1pt}}}%
\hfill\frame{\parbox{0.5666\linewidth}{\centering%
\includegraphics[width=0.5\linewidth]{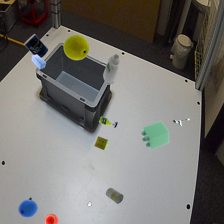}%
\includegraphics[width=0.5\linewidth]{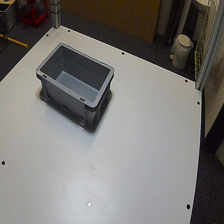}\\\nointerlineskip
\includegraphics[width=0.5\linewidth]{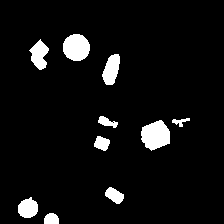}%
\includegraphics[width=0.5\linewidth]{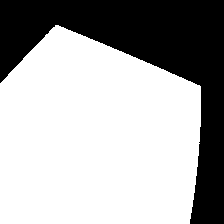}}}%
\vspace*{-2.5ex}
\caption{\textbf{Left:} Easy, medium, and hard examples (left to right) of the \emph{jello}, \emph{minibus}, \emph{bowl}, and \emph{green pepper} test classes (top to bottom). \textbf{Right:} Example masked autoencoder input (top left), target reconstruction (top right), target anomaly mask (bottom left), and target table mask (bottom right), for the training dataset with distractor objects.}
\figlabel{test_object_crops}
\vspace*{-4ex}
\end{figure}

The training loss consists of three terms, an RGB reconstruction loss $L_r$, a table mask loss $L_t$, and an anomaly mask loss $L_a$. The reconstruction term $L_r$ is the mean squared error over the full RGB output, weighted by the target table and anomaly masks so that only table pixels contribute, with anomalous table pixels being weighted 6 times higher than non-anomalous table pixels. This encourages faithful reconstructions focusing on the table area and the removal of pasted objects. The table mask loss $L_t$ is the mean binary cross-entropy (BCE) between the predicted table mask and its target. The anomaly mask loss $L_a$ is similarly defined, but with a weight of 11 applied to anomalous pixels in order to mitigate the class imbalance. Unlike \cite{MaskedAutoencoders2021}, all three loss terms are computed over the entire image, and not only on the masked patches. The final training loss $L$ is
\begin{equation}
L = w_r L_r + w_t L_t + L_a,
\end{equation}
where $w_r = 10$ and $w_t = 2$. The modified MAE is trained for 2000 epochs with batch size 192 using AdamW with a weight decay of 0.05, and a cosine decay schedule with 40 warmup epochs and a base learning rate of 0.0006.

%===============================================================================
\vspace{-2.5ex}
\subsection{Anomaly Localization}
\seclabel{anomaly_localization}
\vspace{-0.5ex}

Given the reconstruction and anomaly mask outputs of the MAE, the goal during inference is to localize individual anomalies and generate class-agnostic bounding boxes for subsequent classification by NOVIC (see \figref{pipeline_overview}). A reconstruction error score map is computed as the pixelwise $L^1$ norm between the predicted $\hat{I}$ and target $I$ RGB values, normalized by a fixed scalar hyperparameter $s_r$, \ie
\begin{equation}
S_r = \mathrm{min} \bigl( \tfrac{1}{s_r} \tilde{M}_t \| \hat{I} - I \|_1, 1 \bigr),
\end{equation}
where $\tilde{M}_t \in \{0, 1\}^{W \times H}$ is the predicted binary table mask (see \secref{masked_autoencoder}), and $\|\!\cdot\!\|_1$ is computed per pixel. Errors outside the predicted table mask are zeroed, as the reconstruction loss $L_r$ only incentivizes accurate reconstruction within the table workspace. Thresholding $S_r$ yields a binary reconstruction error mask, whose connected components above a minimum area are treated as \emph{reconstruction-based anomalies} that are individually scored by the mean of the lower half of their respective $S_r$ pixel scores. The same localization and scoring procedure is applied to the predicted binary anomaly mask (see \secref{masked_autoencoder}) to also obtain the \emph{mask-based anomalies}. The resulting mask-based and reconstruction-based anomalies are then merged and deduplicated to obtain the final class-agnostic bounding boxes passed to NOVIC. If a pair of anomalies has a bounding box IoU above 0.65, only the higher-scoring anomaly is kept. Additional filtering removes anomalies that significantly intersect multiple anomalies of the other type. In both cases, the remaining anomalies have their scores adjusted based on their maximum bounding box or segmentation IoU with an anomaly of the other type, down to a minimum threshold. This rewards anomalies supported by both sources, particularly those with strongly corroborated segmentations, and combines the best of both modalities---the reconstruction mask tends to struggle with gray-white objects that the anomaly mask detects well, while the anomaly mask tends to fuse objects that the reconstruction mask can separate. Optionally, SAM 2.1 \cite{ravi2024sam2segmentimages} is used to refine the detected bounding boxes for more precise localization, turning AnomNOVIC into AnomNOVIC-S. SAM can optionally further be used via targeted point prompts to break up any still-fused anomalies into separate detections, but due to the added runtime overhead of an extra invocation this is disabled by default. The full anomaly localization process is visualized in \figref{anomaly_loc_example}.

\begin{figure}[!t]
\centering%
\parbox{0.5\linewidth}{\centering%
\includegraphics[width=0.3333\linewidth, height=0.3333\linewidth]{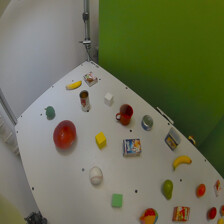}%
\includegraphics[width=0.3333\linewidth, height=0.3333\linewidth]{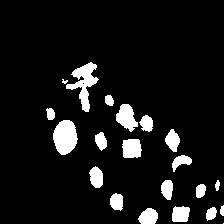}%
\includegraphics[width=0.3333\linewidth, height=0.3333\linewidth]{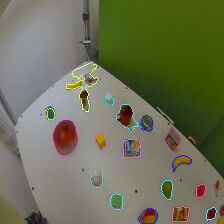}\\\nointerlineskip
\includegraphics[width=0.3333\linewidth, height=0.3333\linewidth]{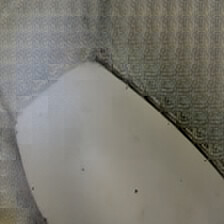}%
\includegraphics[width=0.3333\linewidth, height=0.3333\linewidth]{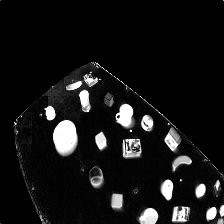}%
\includegraphics[width=0.3333\linewidth, height=0.3333\linewidth]{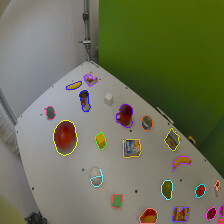}}%
\parbox{0.5\linewidth}{\centering\includegraphics[width=1.0\linewidth, trim=0px 0px 0px 53px, clip]{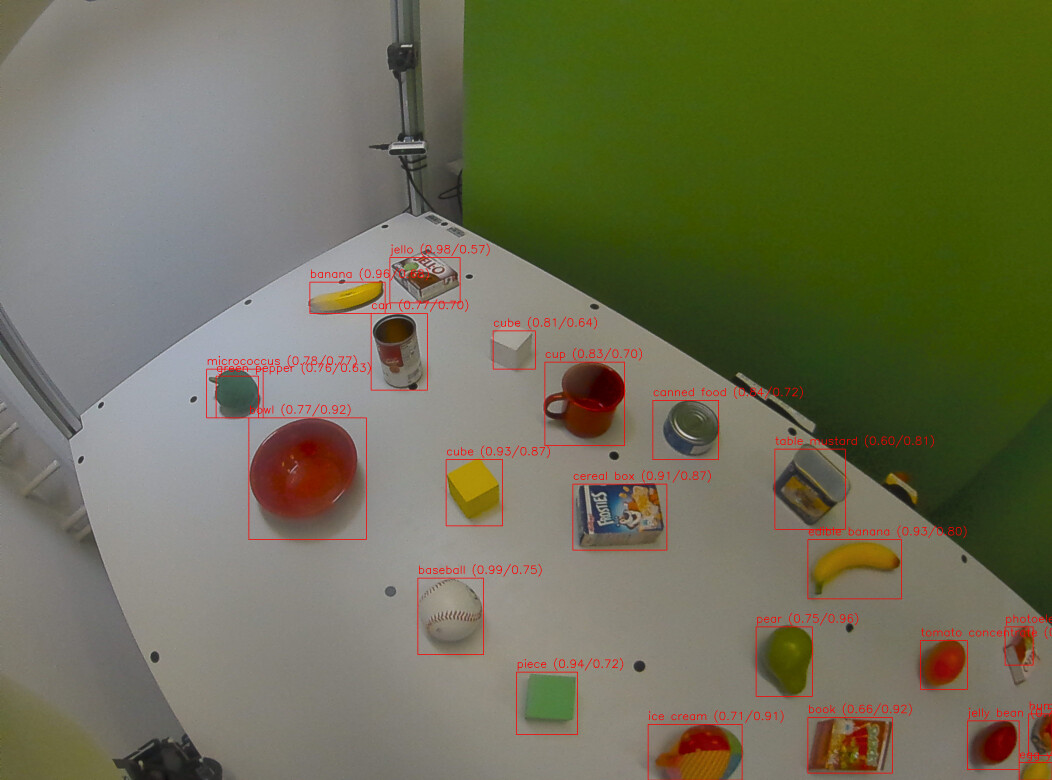}}%
\vspace*{-1.5ex}
\caption{Anomaly localization example, showing the input image, binary anomaly mask, mask-based anomalies, reconstructed image $\hat{I}$, reconstruction error score map $S_r$, reconstruction-based anomalies, and the final merged anomalies with corresponding prompt-free NOVIC classifications and NOVIC\,/\,merged MAE anomaly scores.}
\figlabel{anomaly_loc_example}
\vspace*{-4ex}
\end{figure}

%===============================================================================
\vspace{-1.5ex}
\subsection{Unconstrained Open Vocabulary Image Classification}
\seclabel{novic}
\vspace{-0.5ex}

Given the class-agnostic bounding boxes from anomaly localization, fine-grained classification is performed using NOVIC \cite{allgeuer_novic_2025}. Using slightly dilated anomaly-centered crops, NOVIC generates free-form class label tokens for each anomaly. Although natively prompt-free, unlike almost all existing classifiers, it can also emulate prompted open vocabulary inference by restricting the allowed token choices at each decoding step. In either case, the sum of tokenwise log-probabilities is rescaled by $s_p = 0.14$ prior to exponentiation, and multiplied by the class-agnostic MAE anomaly confidence to obtain the final anomaly score.

NOVIC is trained on a large-scale synthetic caption-object dataset generated from the object noun dictionary \cite{allgeuer_novic_2025} and its noun frequency metadata. Frequency thresholds can be used to eliminate very uncommon nouns, yielding less noisy models. FT2 denotes an applied frequency threshold of 2, while FT0 implies use of the full dataset. Manually modifying noun frequencies reweights noun prevalence in training and changes which nouns are dropped first under thresholding. In this work, we exploit this and reassign noun frequencies based on the likelihood that each object noun could be encountered by the NICOL in its laboratory environment. Using the \texttt{gpt\_batch\_api} library \cite{allgeuer_gpt_2025} and OpenAI's GPT-4o mini LLM, we perform this reassignment for all 42919 object nouns. We then train new FT0 and FT2 NOVIC variants, \eg N-FT0, and compare them to the pretrained FT0 and FT2 models. All four use the DFN-5B H/14-378 CLIP model \cite{fang2024data}.

%%%%%%%%%%%%%%%%%%%%%%%%%%%%%%%%%%%%%%%%%%%%%%%%%%%%%%%%%%%%%%%%%%%%%%%%%%%%%%%%
\vspace{-2.5ex}
\section{Experiments and Evaluation}
\seclabel{experiments_evaluation}
\vspace{-2ex}

The results of the prompt-free AnomNOVIC anomaly recognition pipeline are shown in \tabref{results} and compared against YOLOE \cite{wang2025yoloerealtimeseeing}, the only available prompt-free baseline, though with a smaller fixed vocabulary compared to NOVIC's broad coverage. Hyperparameters were coarsely tuned without overspecialization by ablation from qualitative defaults. AnomNOVIC outperforms YOLOE by a large margin, more than doubling the best YOLOE model's AP$_{50}$ on the plain dataset, with particularly strong AnomNOVIC-S AP scores. Overall, AnomNOVIC shows stronger performance in object detection and classification than in fine-grained bounding box localization. This is reflected in the larger gap between AP and AP$_{50}$, and also explains why applying SAM significantly improves AP, despite SAM not contributing to initial RoI detection. Some localization imprecision is expected, however, since the MAE operates at only 224px. Given the small object sizes, a bounding box error of just one pixel on all four sides is sufficient to miss all AP categories from 75 to 95. Nevertheless, the \textit{class-agnostic} anomaly detection performance of AnomNOVIC remains high, reaching 49.7\% AP and 88.2\% AP$_{50}$ on the plain dataset, or 69.5\% and 91.2\% with SAM. On the distractors dataset, the corresponding values are 51.4\% and 84.4\% without SAM, and 71.9\% and 87.0\% with SAM. The non-SAM runtime is $\approx$\,500\,ms end-to-end.

\begin{table}[t]
    \centering\footnotesize
    \caption{Anomaly recognition results (\%) for the plain ($p$), distractor ($d$), and wild ($w$) datasets. The best two results per column are \textbf{bold}, the best baseline is \textit{italicized}.}
    \tablabel{results}
    \vspace*{-2ex}
    \begin{tabular}{@{\hspace{0.5\tabcolsep}}l@{\hspace{10pt}}ccccc@{\hspace{10pt}}cccc@{\hspace{0.5\tabcolsep}}}
    \toprule
    & \multicolumn{5}{c}{\textbf{Prompt-free}} & \multicolumn{4}{c}{\textbf{Prompted}} \\
    \textbf{Model} & \textbf{Acc$^w$} & \textbf{AP$^p$} & \textbf{AP$_{50}^p$} & \textbf{AP$^d$} & \textbf{AP$_{50}^d$} & \textbf{AP$^p$} & \textbf{AP$_{50}^p$} & \textbf{AP$^d$} & \textbf{AP$_{50}^d$} \\
    \midrule
    YOLOE-11M & 10.5 & 18.3 & 21.0 & 21.9 & 24.7 & 21.6 & 24.9 & 25.1 & 28.4 \\
    YOLOE-11L & \textit{14.2} & \textit{20.4} & \textit{23.7} & \textit{30.8} & \textit{34.7} & 19.5 & 22.5 & 28.4 & 31.8 \\
    OWLv2 CLIP B/16 & - & - & - & - & - & 19.1 & 24.2 & \textit{33.4} & \textit{42.9} \\
    OWLv2 CLIP L/14 & - & - & - & - & - & 18.8 & 22.9 & 32.1 & 37.9 \\
    YOLO-World-v2-M & - & - & - & - & - & 18.6 & 20.0 & 21.7 & 24.5 \\
    YOLO-World-v2-L & - & - & - & - & - & 23.3 & 24.2 & 27.9 & 30.7 \\
    YOLO-World-v2-X & - & - & - & - & - & \emph{26.0} & \emph{27.8} & 26.5 & 28.9 \\
    \midrule
    AnomNOVIC FT0 & - & 25.7 & 44.7 & 31.9 & 53.2 & 35.0 & 61.2 & 38.9 & 64.4 \\
    AnomNOVIC FT2 & - & 26.5 & 45.4 & 33.6 & 55.7 & 36.8 & 64.4 & 41.1 & 68.2 \\
    AnomNOVIC N-FT0 & - & 24.9 & 42.9 & 30.0 & 50.7 & 36.5 & 63.8 & 39.9 & 66.5 \\
    AnomNOVIC N-FT2 & - & 28.2 & \textbf{48.9} & 33.3 & 55.8 & 36.3 & 63.7 & 40.6 & 67.5 \\
    \midrule
    AnomNOVIC-S FT0 & 80.3 & 35.1 & 45.5 & 44.2 & 54.6 & 48.9 & 63.6 & 56.0 & 69.3 \\
    AnomNOVIC-S FT2 & 76.6 & \textbf{36.1} & 46.6 & \textbf{47.1} & \textbf{57.5} & \textbf{51.1} & \textbf{66.7} & \textbf{59.0} & \textbf{72.5} \\
    AnomNOVIC-S N-FT0 & \textbf{82.6} & 34.0 & 44.0 & 42.1 & 52.0 & \textbf{50.8} & \textbf{66.2} & 57.5 & 70.9 \\
    AnomNOVIC-S N-FT2 & \textbf{82.1} & \textbf{38.3} & \textbf{49.8} & \textbf{46.9} & \textbf{57.7} & 50.6 & 66.1 & \textbf{58.2} & \textbf{71.9} \\
    \bottomrule
    \end{tabular}
    \vspace*{-4ex}
\end{table}

AnomNOVIC exhibits strong open vocabulary classification ability. On the Wild dataset, which contains very diverse and challenging objects, it achieves open-set detection and classification accuracies of up to Acc$^w = 82.6\%$, compared to the best state-of-the-art YOLOE accuracy of 14.2\% (see \tabref{results}). Highest-scoring detections above an IoU of 50\% were considered using correct-close-incorrect scoring \cite{allgeuer_novic_2025}. Due to the high manual annotation cost of open-set recognition, not all AnomNOVIC variants were evaluated on this dataset. Qualitative examples of detections for YOLOE and AnomNOVIC are shown in \figref{wild_comparison}. AnomNOVIC frequently yields fine-grained correct classifications such as \emph{silver medal}, \emph{fine tooth comb}, \emph{audio amplifier}, and \emph{electronic voltmeter}, whereas YOLOE produces many high-confidence false positives, limiting its reliability and practical utility for most downstream applications. The evaluation of AnomNOVIC in a prompted setting allows comparison with additional state-of-the-art open vocabulary methods, including OWLv2 \cite{minderer2023scaling} and YOLO-World-v2 \cite{Cheng2024YOLOWorld}. As shown in \tabref{results}, AnomNOVIC again more than doubles the performance of the strongest competing model on the plain dataset, with trends similar to the prompt-free setting occurring in terms of AP versus AP$_{50}$ and the impact of SAM, also for the distractors dataset. Overall, the results indicate that retraining NOVIC produced a more effective detection model for the NICOL. In the prompt-free setting, N-FT2 achieved the best detection performance. However, on the Wild dataset, N-FT0 achieved even higher classification accuracy, suggesting that the full unthresholded training dataset is particularly beneficial in more exotic open-set scenarios. The Wild dataset performance drop from N-FT0 to N-FT2 is however much smaller than from FT0 to FT2, highlighting that reweighting the NOVIC training dataset helped prune less relevant object nouns under frequency thresholding.

AnomNOVIC can also operate in more complex environments that require a freely moving camera perspective, such as a mobile robot navigating within fixed room(s), provided that the range of normal workspace variation is known and included in the training data. To demonstrate this, we collected training data from the interior of an airplane fuselage as well as some test images containing anomalous objects, and applied prompt-free AnomNOVIC to detect them (see \figref{wild_comparison}). The shown test images were captured from novel viewpoints not present in the training data, indicating generalization across viewpoint changes within a fixed environment. In this case, the `table mask' was adapted to exclude regions outside the fuselage that are visible through open windows and doors.

\begin{figure}[!t]
\parbox{\linewidth}{\centering%
\includegraphics[width=0.5\linewidth, trim=0px 0px 0px 60px, clip]{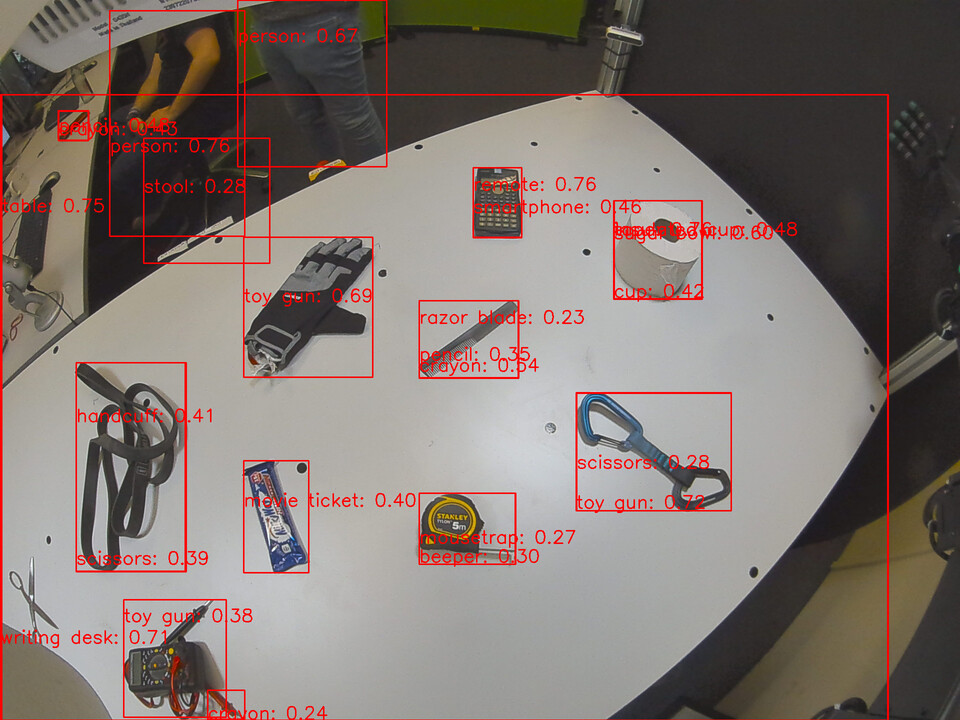}%
\includegraphics[width=0.5\linewidth, trim=0px 0px 0px 60px, clip]{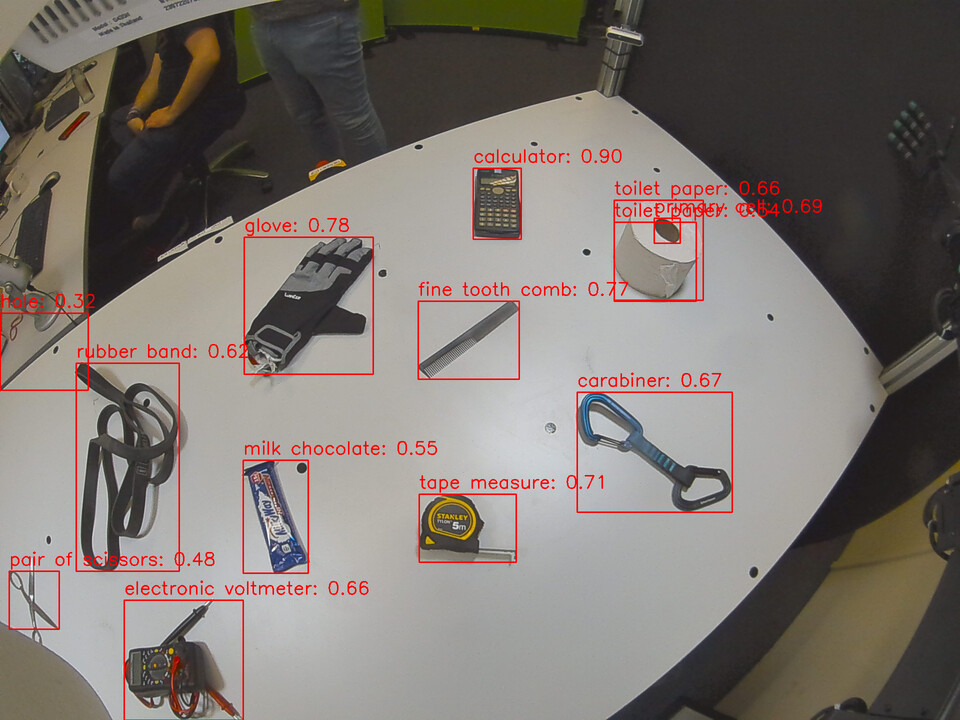}\\\nointerlineskip
\includegraphics[width=0.5\linewidth]{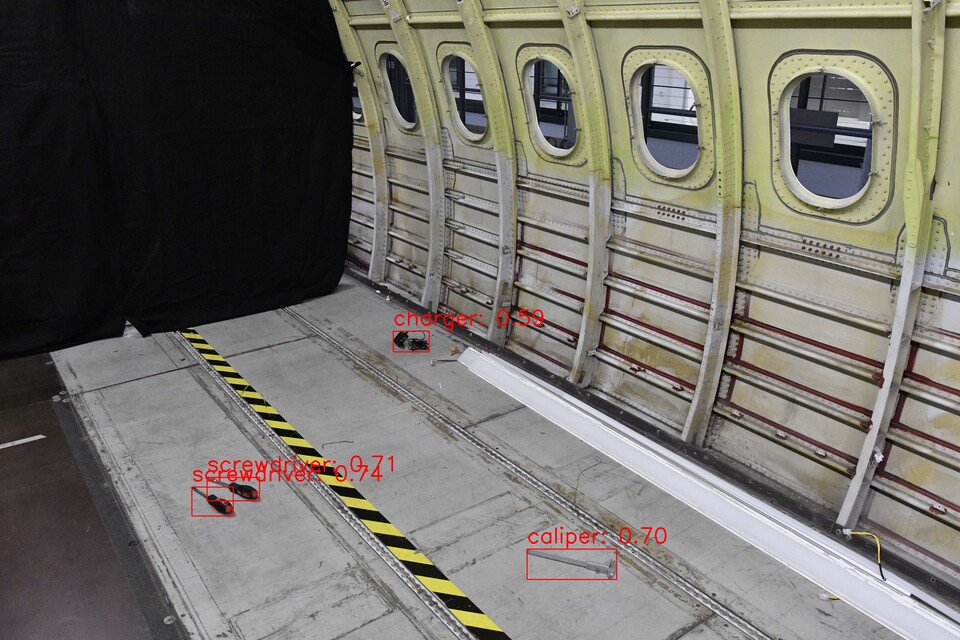}%
\includegraphics[width=0.5\linewidth]{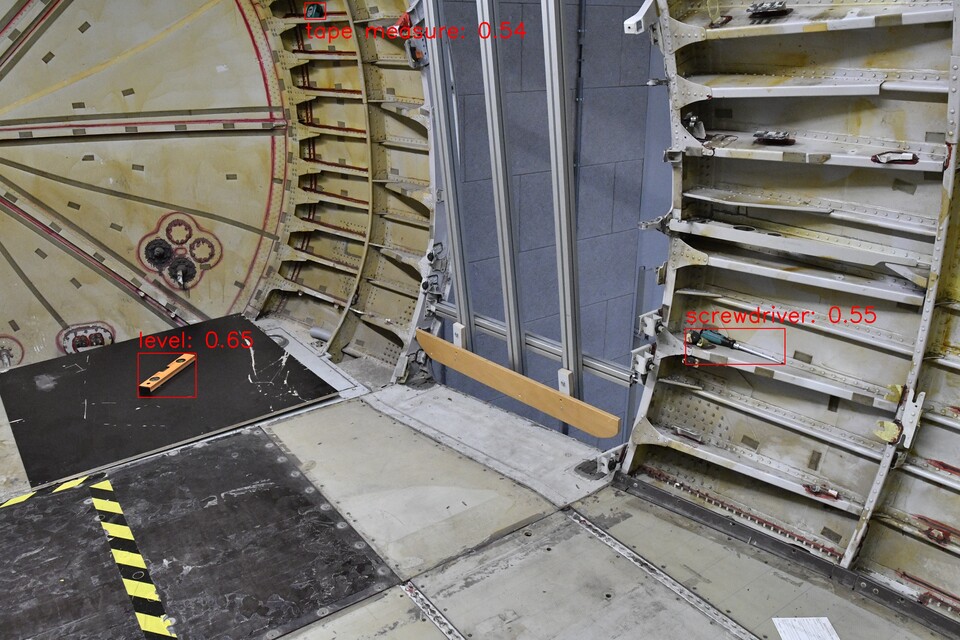}}%
\vspace*{-2ex}
\caption{\textbf{Top:} Prompt-free predictions of YOLOE-11L (left) and AnomNOVIC-S N-FT0 (right) on the Wild dataset, showing the starkly better performance of AnomNOVIC. \textbf{Bottom:} Prompt-free predictions with a moving camera in a fuselage environment.}
\figlabel{wild_comparison}
\vspace*{-3ex}
\end{figure}

%%%%%%%%%%%%%%%%%%%%%%%%%%%%%%%%%%%%%%%%%%%%%%%%%%%%%%%%%%%%%%%%%%%%%%%%%%%%%%%%
\vspace{-2ex}
\section{Conclusion}
\seclabel{conclusion}
\vspace{-2ex}

We presented AnomNOVIC, a prompt-free open vocabulary recognition system for known workspaces. Our approach contributes to both anomaly detection and NOVIC model tuning, combining class-agnostic MAE-based object localization with unconstrained open vocabulary image classification, without requiring class prompts or task-specific finetuning. Experiments on controlled and in-the-wild datasets demonstrate state-of-the-art performance, advancing the goal of semantically aware robots that can perceive and interact with open-ended environments.

% ---- Bibliography ----
%
% BibTeX users should specify bibliography style 'splncs04'.
% References will then be sorted and formatted in the correct style.
%
\bibliographystyle{splncs04}
{%
\let\oldthebibliography\thebibliography%
\renewcommand{\thebibliography}[1]{\vspace{-3ex}\oldthebibliography{#1}\vspace{-2ex}}%
\bibliography{IEEEabrv, main}%
}

\end{document}